\documentclass[lettersize,journal]{IEEEtran}
\usepackage{amsmath,amsfonts}
\usepackage{algorithmic}
\usepackage{array}
\usepackage{textcomp}
\usepackage{stfloats}
\usepackage{url}
\usepackage{verbatim}
\usepackage{graphicx}

\usepackage{algorithm}
\usepackage{xcolor}

\usepackage{physics}
\usepackage{mathdots}
\usepackage{amssymb}
\usepackage{subfigure}
\usepackage{pgfplots}
\usepackage{tikz}
\usetikzlibrary{patterns}
\usepackage{multirow}
\usepackage{multicol}
\usepackage{authblk}
\usepackage{array}
\usepackage{makecell}

\usepackage[all]{nowidow}

\usepackage{hyperref}
\hypersetup{
    colorlinks=true,
    linkcolor=blue,
    filecolor=magenta,      
    urlcolor=cyan,
    pdftitle={Overleaf Example},
    pdfpagemode=FullScreen,
    }

\def\BibTeX{{\rm B\kern-.05em{\sc i\kern-.025em b}\kern-.08em
    T\kern-.1667em\lower.7ex\hbox{E}\kern-.125emX}}
\usepackage{balance}
\begin{document}

\title{Zero-Shot Cellular Trajectory Map Matching}

\author{Weijie Shi, Yue Cui, Hao Chen, Jiaming Li, Mengze Li, Jia Zhu, Jiajie Xu, and Xiaofang Zhou
\thanks{Manuscript created March, 2025; This work was developed by the Soochow University. (\textit{Corresponding author: Jiajie Xu.})}
\thanks{Weijie Shi and Jiajie Xu are with Soochow University, Suzhou 215006, China (e-mail: shiweijie0311@foxmail.com; xujj@suda.edu.cn).}
\thanks{Yue Cui and Xiaofang Zhou are with Hong Kong University of Science and Technology, Hong Kong 999077, China (e-mail: ycuias@cse.ust.hk; zxf@cse.ust.hk).}
\thanks{Hao Chen is with Tencent Inc., Shenzhen 518054, China (e-mail: herrickchen@tencent.com).}
\thanks{Jiaming Li is with ByteDance Inc., Hangzhou 311121, China (e-mail: lijiaming.1124@bytedance.com).}
\thanks{Mengze Li is with Zhejiang University, Hangzhou 310058, China (e-mail: mengzeli@zju.edu.cn).}
\thanks{Jia Zhu is with Zhejiang Normal University, Jinhua 321004, China (e-mail: jiazhu@zjnu.edu.cn).}
}

\markboth{Journal of \LaTeX\ Class Files,~Vol.~18, No.~9, September~2020}%
{How to Use the IEEEtran \LaTeX \ Templates}

\maketitle

\begin{abstract}
Cellular Trajectory Map-Matching (CTMM) aims to align cellular location sequences to road networks, which is a necessary preprocessing in location-based services on web platforms like Google Maps, including navigation and route optimization. Current approaches mainly rely on ID-based features and region-specific data to learn correlations between cell towers and roads, limiting their adaptability to unexplored areas. To enable high-accuracy CTMM without additional training in target regions, Zero-shot CTMM requires to extract not only region-adaptive features, but also sequential and location uncertainty to alleviate positioning errors in cellular data. In this paper, we propose a pixel-based trajectory calibration assistant for zero-shot CTMM, which takes advantage of transferable geospatial knowledge to calibrate pixelated trajectory, and then guide the path-finding process at the road network level. To enhance knowledge sharing across similar regions, a Gaussian mixture model is incorporated into VAE, enabling the identification of scenario-adaptive experts through soft clustering. To mitigate high positioning errors, a spatial-temporal awareness module is designed to capture sequential features and location uncertainty, thereby facilitating the inference of approximate user positions. Finally, a constrained path-finding algorithm is employed to reconstruct the road ID sequence, ensuring topological validity within the road network. This process is guided by the calibrated trajectory while optimizing for the shortest feasible path, thus minimizing unnecessary detours. Extensive experiments demonstrate that our model outperforms existing methods in zero-shot CTMM by 16.8\%.
\end{abstract}


\section{Introduction}
Cellular trajectory data refers to location sequences recorded by cell towers associated with mobile phones, continuously and passively collected on a massive scale of rich information about people's movements. The widespread of large-scale cellular trajectory datasets has made them indispensable for various location-based applications on web platforms like Google Maps \cite{ritchie2024google} and Waze \cite{laor2022waze}, including navigation \cite{chen2007geotracker}, traffic scheduling \cite{zhao2024trident}, route optimization \cite{ma2024more,gao2024lightweight}, and human mobility analysis \cite{8014487,8737535,wang2024cola}. To prepare these data for downstream tasks, a key preprocessing step is Cellular Trajectory Map-Matching (CTMM), which aligns the cellular trajectories onto road networks, restoring the paths traveled by users. 

Unfortunately, compared to high-precision GPS trajectories (with errors of 1-50 meters), cellular trajectories have much higher positioning errors (100-3000 meters), making traditional methods \cite{ST-matching,ivmm,If-matching} struggle to accurately restore user location based on explicit features and hand-designed heuristics. To mitigate the effects of high location uncertainty, recent learning-based methods like sequence-to-sequence models \cite{DMM,DeepMM} and neuralized HMMs \cite{LHMM}, use ID-based features to model implicit trajectory-road associations. Nonetheless, ID-based features heavily rely on statistical patterns between specific cell towers and roads within the training data, leading to mismatches in unobserved regions. While annotation data is extremely scarce due to privacy policies, there is a pressing need for zero-shot CTMM capable of generalizing beyond region-specific data. 

To enable zero-shot CTMM, it is crucial to perform map-matching on a shared geospatial representation. Inspired by TrjSR \cite{cao2021accurate}, we employ a pixel-based grid map to depict both trajectories and road networks within a shared space. In such way, cellular trajectories and road networks can be transformed into calibrated trajectories at the pixel-level, facilitating accurate user localization. To obtain topologically valid road ID sequences, a path-finding algorithm can be guided by these calibrated trajectories. However, achieving both high accuracy and transferability is not trivial, it presents several significant challenges:
\begin{itemize}
    \item \textbf{How to enhance the region-adaptive ability.}
    Moving patterns can vary greatly across different regions, in aspects like road network structure, road density, and placement of cell towers. To adapt to these variations in a zero-shot setting, it would be beneficial to customize experts with region-specific knowledge and share knowledge across similar regions, so that the model can quickly adapt to new regions by leveraging relevant knowledge from similar regions it has encountered before.

    \item \textbf{How to mitigate high location uncertainty.} 
    The extensive coverage area of cell towers results in a single tower potentially associating with multiple roads, significantly increasing location uncertainty. To address this, explicitly predicting uncertainty-related characteristics between cell towers and user location allows the model to approximate likely paths. Additionally, extracting temporal features like direction also helps eliminate improbable positions, further refining location estimates.
    
    \item \textbf{How to robustly convert calibrated trajectory to road-level path.} 
    The ultimate output required for CTMM is a sequence of road IDs, this necessitating the seamless integration of a path-finding algorithm to transform calibrated trajectory from pixel-level into road-level paths. Such a path-finding algorithm should utilize calibrated trajectories as guidance while considering topological accessibility, turn constraints at junctions, and shortest-path preferences to avoid matching detours.
\end{itemize}

In this paper, we propose ZSMM, a two-stage framework for zero-shot map-matching of cellular trajectories. It first calibrates cellular trajectory at the shared pixel space, which then guides the search for a resulting path at the road level. Specifically, ZSMM first pixelates the cellular trajectory and road network into image formats, and then a variational autoencoder (VAE) backbone converts them into a calibrated trajectory image. To share knowledge among similar regions, a Gaussian mixture model (GMM) is integrated into the VAE, where each component is an expert tailored to a type of map-matching scenario. By maximizing the evidence lower bound via stochastic gradient variational Bayes (SGVB), distinctive regions can be separated into different GMM components, enabling dynamic selection of region-adaptive experts. To overcome high location uncertainty, a spatial-aware network empower VAE by explicitly charactering uncertainty-related factors like distance and variance, and a temporal-aware network captures sequential features to eliminate improbable paths. Finally, to ensure topological validity at the road level, a constrained path-finding algorithm is designed to search the optimal path under trajectory calibration constraints while minimizing path length to avoid detours. Extensive experiments demonstrate state-of-the-art zero-shot performance on real-world and synthetic datasets, proving ZSMM's effectiveness and transferability. The main contributions are summarized as follows:
\begin{itemize}
    \item We are the first to propose a pixel-based trajectory calibration assistant for zero-shot CTMM, followed by constraint path-finding to result in road sequence. Once trained, the model can effectively transfer across unseen regions without any extra training.    
    \item We incorporate a GMM into VAE to enhance region-adaptive ability by knowledge sharing among similar regions while customizing for distinct areas.
    \item We design a spatial-aware network to explicitly characterize location uncertainty, and a temporal-aware network to capture sequential features, jointly mitigating the high uncertainty inherent in cellular trajectories.
    \item We design a constrained path-finding algorithm that takes the calibrated trajectory as guidance while minimizing path length to avoid detours.
\end{itemize}

\section{RELATED WORK}
Map-matching is a critical preprocessing step for location-based web applications, aiming to align trajectory points with road networks to identify the path traveled by a vehicle or user. Compared to GPS trajectories, cellular trajectories have larger-scale data, as cell towers passively collect user mobility information at all times \cite{rizk2019solocell}. Unfortunately, they also have much higher positioning errors, posing tremendous challenges for map-matching \cite{yang2010accuracy}. A variety of map-matching algorithms have been proposed for cellular trajectories, which can be categorized into traditional methods and learning-based methods \cite{chao2020survey,huang2021survey}.

\textbf{Traditional CTMM Methods.}
Intuitively, similarity models \cite{mosig2005approximately,joshi2001new,brakatsoulas2005map} identify the geometrically and/or topologically closest vertices/edges to the trajectory, but they lack a global perspective. In contrast, score-based models \cite{mm-survey} incorporate global views by evaluating road candidates and selecting the one with the maximum score computed by an objective function. In particular, Hidden Markov Model (HMM)-based methods \cite{8315392,ST-matching,ivmm} have proven particularly effective, modeling observation and transition probabilities based on distance and road connectivity. On top of this, many works \cite{Clsters,mcm} have developed various heuristics to address ambiguous cases and location uncertainty, such as mining speed information \cite{If-matching}, applying filters to mitigate positioning errors \cite{Snapnet}, introducing global weighted-matrix voting \cite{zhang2021rcivmm}, and using geometric relationships between road segments as constraints \cite{THMM}. Despite this, these approaches rely on explicit features and hand-designed heuristics, which are inadequate for cellular trajectory data with larger positioning errors, necessitating more robust similarity criteria between trajectories and paths.

\begin{figure*}[t]
\centering
\begin{minipage}[t]{1\linewidth}
\centering
\includegraphics[width=6.8in]{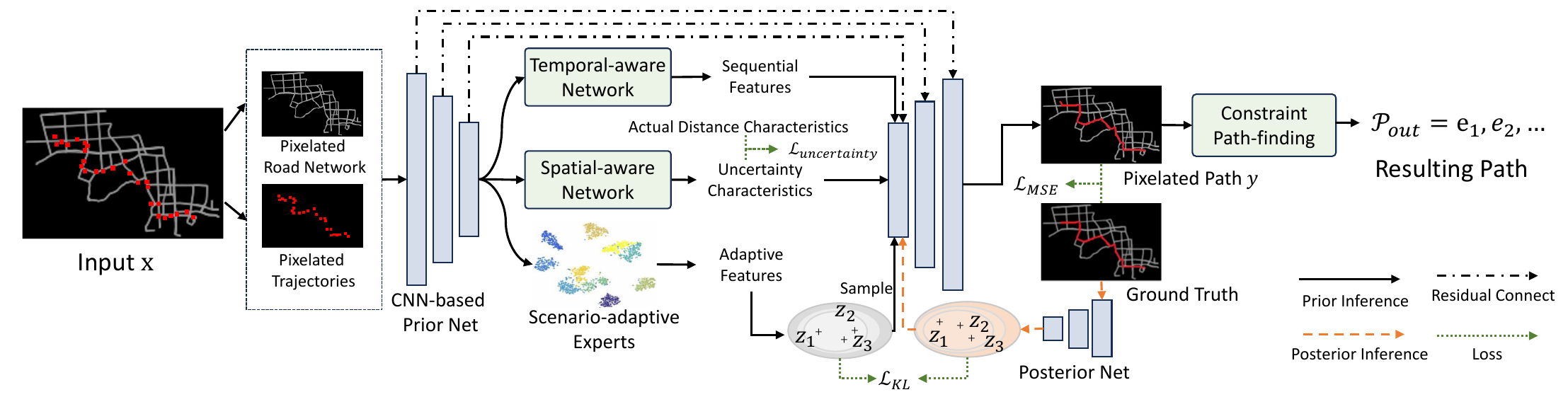}
\end{minipage}
\centering
\caption{The overview of ZSMM, including 1) Pixelization module for transforming input data into a shared pixel space; 2) Trajectory calibration module with a VAE backbone, incorporating spatial-aware and temporal-aware networks to handle location uncertainty and sequential features; 3) Scenario-adaptive experts using GMM for region-adaptive learning; 4) Constraint path-finding module to generate the final road-level path based on the calibrated trajectory.}
\label{fig:framework}
\end{figure*}
\textbf{Learning-Based CTMM Methods.}
Recent advancements in deep learning have facilitated the development of learning-based CTMM methods \cite{karatzoglou2018seq2seq,hashemi2016machine,liang2016online,liu2020deep}, which effectively capture complex relationships between trajectories and road networks. Approaches such as DeepMM \cite{DeepMM}, DMM \cite{DMM}, TransformerMM \cite{transformerMM}, and L2MM \cite{L2mm} utilize neural networks like RNNs and Transformers to learn generative models that map cell tower ID sequences to road ID sequences. GraphMM \cite{liu2023graphmm} leverages graph neural networks to capture inter-trajectory and trajectory-road correlations, enhancing prediction accuracy and efficiency. LHMM \cite{LHMM} neuralizes HMM to automatically extract correlations between cell towers and roads by considering trajectory contexts, reducing the need for hand-designed heuristics. However, these learning-based models predominantly focus on obtaining high-quality tower and road ID embeddings, requiring large amounts of labeled data and limiting adaptability to unobserved regions. To enhance transferability, we propose a shift from ID-based to pixel-based models. This approach represents geospatial data in pixel space, enabling zero-shot map-matching in new regions without retraining.

\section{PROBLEM FORMULATION}
We first define some key concepts of CTMM, and then formalize the objective of this paper.

\textbf{Definition 1 (Cell Tower Sample).} A cell tower is characterized by a fixed spatial position, represented by a longitude and latitude coordinate pair: $pos=(lon,lat)$. When a mobile phone establishes communication with a cell tower (e.g. network service requests), a cell tower sample is formed, denoted as $x = (user_{ID}, pos, t)$, where $user_{ID}$ is an anonymized user identifier and $t$ is the timestamp of the communication event.

\textbf{Definition 2 (Cellular Trajectory).} As a user goes on a trip, their mobile phone continuously communicates with surrounding cell towers, generating a series of cell tower samples: $\mathcal{T}=p_1, p_2, \dots, p_{|\mathcal{T}|}$, named as a cellular trajectory.

\textbf{Definition 3 (Road Network).} A road network can be represented as $G(V,E)$, where $V$ denotes a set of nodes corresponding to intersections or terminal points, and $E$ contains a set of directed road segments $e_i$ connecting these nodes.

\textbf{Definition 4 (Path).} A path is defined as a sequence of road segments, expressed as $\mathcal{P} = e_1, e_2, \dots, e_{|\mathcal{P}|}$, where $e_i$ is a road segment on top of road network $G$, and the ending node of $e_i$ coincides with the starting node of $e_{i+1}$.

\textbf{Problem Formalization:} Given a road network $G$ and a cellular trajectory $\mathcal{T}$, the objective of the cellular trajectory map-matching problem is to accurately match a path $\mathcal{P}_{out}$ that closely approximates the ground-truth path $\mathcal{P}_g$ corresponding to $\mathcal{T}$.

\section{METHODOLOGY}
In this section, we present details of our model, including an overview and key components.

\subsection{Overview}
The overview of ZSMM is illustrated in Figure \ref{fig:framework}. It includes a trajectory calibration process to locate the approximate geographic position of the user in pixel space, which further guides the road-level path-finding process. 

Specifically, the pixelization module first transforms the input cellular trajectory and road network into a shared pixel space. Next, the pixelated trajectory calibration module leverages a VAE as the backbone to generate a calibrated trajectory image. Within the VAE, a spatial-aware network explicitly characterizes location uncertainty, and a temporal-aware network captures sequential features, jointly mitigating the high uncertainty inherent in cellular trajectories. To enhance region-adaptive ability, a GMM is incorporated into the VAE, enabling knowledge sharing among similar regions while customizing for distinct areas. Finally, a constrained path-finding module takes the calibrated pixel trajectory as guidance to find the optimal road-level path. This module ensures topological validity within the road network while minimizing path length to avoid unnecessary detours. In the following subsections, we elaborate on the technical details of each module.

\subsection{Pixelated Trajectory Calibration}
\begin{figure*}[t]
\centering
\begin{minipage}[t]{1\linewidth}
\centering
\includegraphics[width=6.8in]{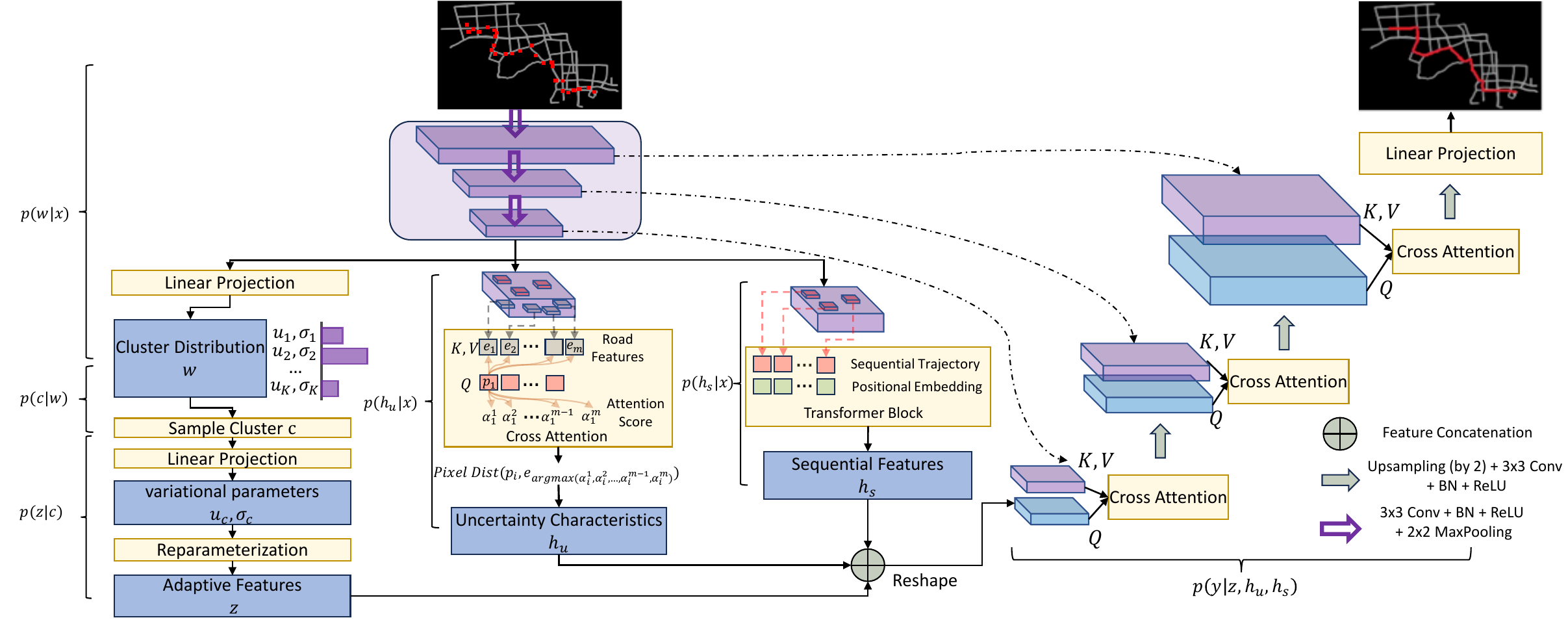}
\end{minipage}
\centering
\caption{The pixelated trajectory calibration. This process begins with a CNN for spatial feature extraction, followed by scenario-adaptive experts using GMM for region-specific learning. It incorporates a spatial-aware network to characterize location uncertainty and a temporal-aware network with transformers to capture sequential features. The decoding stage employs cross-attention and upsampling operations to generate the calibrated trajectory.}
\label{fig:map-matcher}
\end{figure*}

To enable zero-shot transferability, we transform the raw inputs into a shared pixel space, and learn matching patterns from pixelated cellular trajectories and road networks to calibrated pixel trajectories as Figure \ref{fig:map-matcher}. To overcome the high positioning uncertainty in cellular data, a temporal-aware network captures sequential features to eliminate improbable paths, while a spatial-aware network explicitly models uncertainty-related characteristics to approximate likely user positions. To enhance region-adaptive ability, we incorporate a GMM into the VAE backbone, allowing for knowledge sharing among similar regions through selecting scenario-adaptive experts. Finally, cross-attentions and upsampling operations are utilized to decode the latent features and locate the potential pixel positions traversed by the user.

\subsubsection{Pixelated Representation}
To represent trajectories, road networks, and paths in pixel space while preserving the underlying spatial and structural information, we convert them into image format. Specifically, we design an orthogonal grid coordinate system to discretize the geographic region into pixel cells. The cellular trajectory with $|\mathcal{T}|$ points is converted into a pixel matrix $\mathcal{T}_{img}$ denoting the pixel cells passed through. For the $i$-th trajectory point $p_i$, its presence in a pixel cell is represented by a value of $\frac{i}{|\mathcal{T}|}$, while absence is denoted by 0. For the pixelated road network $G_{img}$, a pixel cell with value 1 indicates the presence of a road segment, while 0 means no road. By stacking $\mathcal{T}_{img}$ and $G_{img}$ along the channel dimension, the resulting multi-channel images embed trajectory positions as well as topological information about road connectivity and coverage density.

\subsubsection{Temporal-aware Network} 
The conversion of trajectories to pixel space results in the loss of sequential information, while this temporal context is vital for understanding movement patterns and predicting likely paths. To address this limitation, a temporal-aware network is designed to combine positional embedding on pixelated trajectory points. Specifically, to capture both local spatial relationships and global temporal patterns, a series of CNNs first extract spatial hidden features for each position in the pixel space. The extracted spatial features corresponding to each trajectory point are then augmented with Rotary Positional Embedding (ROPE) \cite{su2024roformer}, integrating the trajectory's sequential order information into the feature representation. Next, transformer layers are employed to model the global temporal dependencies, effectively reintroducing the sequential nature of the trajectory data into the pixel-based representation. The temporal-aware network can be formally defined as:
\begin{equation}
    h_s = \text{Transformer}(\text{CNN}(X), \text{RoPE}(p_1, p_2, \ldots, p_{|\mathcal{T}|}))
\end{equation}
where $x$ represents the pixelated input, $p_i$ denotes the i-th trajectory point, $|T|$ is the total number of trajectory points, and $h_s$ is the sequential features.

\subsubsection{Spatial-aware Network}
Cellular trajectory data suffers from high positioning uncertainty due to the large coverage area of cell towers, leading to significant location uncertainty. To mitigate this, a spatial-aware network explicitly accounts for uncertainty-related factors, enabling a more accurate estimation of the user's probable position. Specifically, it operates by characterizing two key aspects for each trajectory point: the pixel distance and variance to its most relevant roads. The relevance of road segments is determined through a cross-attention layer, which computes attention scores between trajectory points and road features. These uncertainty-related factors are helpful in locating approximate user positions after tuning by actual distance characteristics. The spatial-aware network can be formally defined as:
\begin{equation}
d_i = \text{PixelDist}(p_i, e_{\text{argmax}(\alpha_i^j)}), i \in {1, 2, ..., |\mathcal{T}|}, j \in {1, 2, ..., |E|}
\end{equation}
\begin{equation}
h_u = [\mu(d_1, ..., d_{|\mathcal{T}|}), \sigma^2(d_1, ..., d_{|\mathcal{T}|})]
\end{equation}
where $d_i$ represents the pixel distance between the i-th trajectory point $p_i$ and its most relevant road segment, $\alpha_i^j$ is the attention score between the i-th trajectory point and the j-th road segment, $|E|$ is the number of road segments within input image, $\text{PixelDist}(\cdot,\cdot)$ calculates the pixel distance, $\mu(\cdot)$ and $\sigma^2(\cdot)$ computes the mean and variance respectively.

\subsubsection{Scenario-adaptive Experts}
Map-matching scenarios can vary significantly across different regions with changed road density, and cell tower placement. A one-size-fits-all approach may struggle to adapt to these diverse scenarios, potentially leading to suboptimal performance in areas with unique characteristics. Adaptive CTMM calls for sharing knowledge across similar regions while customization for distinct areas. To achieve this balance between knowledge sharing and specialization, we incorporate a GMM within our VAE framework to automatically cluster similar scenarios and develop specialized experts for each cluster.

Specifically, we model the latent space of VAE using a Gaussian mixture distribution with $K$ experts, each representing a distinct map-matching scenario or expert. The process begins with the computation of a Cluster Distribution $w$ through a linear projection of the CNN features: $w = \text{MLP}(\text{CNN}(x))$. This distribution $w$ encodes the relevance of each expert to the current input. From this distribution, we sample top-$k$ experts $c$ from categorical distribution and extract the corresponding hidden features $h_{c_i}$, denoted as $Cat(c|w)$. Next, we generate the variational parameters $\mu_{c_i}$ and $\sigma_{c_i}$ for each sampled expert by a linear projector, followed by reparameterization \cite{kingma2013auto} to obtain adaptive variational variable $z$:
\begin{equation}
z = \sum_{c=1}^{k}{\omega_c(\mu_c+ \sigma_c \odot \epsilon)}
\end{equation}
where $\epsilon \sim \mathcal{N}(0,I)$, $\odot$ represents element-wise multiplication, and ${\omega_c}$ is the weight for cluster $c$. This integrated $z$ encapsulates the characteristics of the chosen experts, enabling the model to tailor its map-matching strategy to the specific scenario at hand.

\subsubsection{Decoding Process}
The decoding process in ZSMM is designed to reconstruct the calibrated trajectory in pixel space while maintaining alignment with the underlying road network. To achieve this, a series of upsampling operations are employed, interleaved with cross-attention layers to guide the hidden features in locating relevant positions within the original road network. Specifically, the sequential features, uncertainty characteristics, and adaptive features are concatenated into a comprehensive initial feature map. This feature map is then progressively refined through a series of upsampling operations aimed at increasing the spatial resolution of the latent representation. Each upsampling stage is followed by a cross-attention layer that aligns the feature map with the original representation of the road network, allowing the model to dynamically attend relevant road features.

\subsubsection{Variational Lower Bound}
To optimize the pixelated map-matcher, we aim to maximize the variational lower bound on the marginal likelihood of the expected pixelated path given cellular trajectory and road network. According to the generative process above, the conditional joint probability $p(y, h_s, h_u, z, c, w| x)$ can be factorized as: 
\begin{equation}
    p(y, h_s, h_u, z, c, w | x) = p(y|z,h)p(h|x)p(z|c)p(c|w)p(w|x)
\end{equation}
where an expected pixelated path $y$ is generated from a set of latent variables $h_s, h_u, z$, cluster $c$, and categorical distribution $w$ conditioned on pixelated input $x$. For simplicity, we denote $h=[h_s, h_u]$ in the following. Based on Jensen’s inequality, the log-likelihood of our model can be written as:
\begin{align}
    \log{p(y|x)}&=\log{\iiint{\sum_{c}{p(y,h,z,c,w|x)}}dhdzdw} \\
    &\geq \mathbb{E}_{q(h,z,c,w|y,x)}[\log{\frac{p(y,h,z,c,w|x)}{q(h,z,c,w|y,x)}}] \\
    &= \mathcal{L}_{\text{ELBO}}(y|x) \label{eq6}
\end{align}
where $\mathcal{L}_{\text{ELBO}}$ is evidence lower bound (ELBO), and $q(h,z,c,w|y,x)$ is a proxy of variational posterior to approximate the true posterior $p(h,z,c,w|y,x)$. In ZSMM, we assume $q(h,z,c,w|y,x)$ to be a mean-field variational family, thereby it can be factorized as:
\begin{equation}
    q(h,z,c,w|y,x) = q(h|y,x)q(z|c)q(c|w)q(w|y,x)
\end{equation}
As depicted in Figure \ref{fig:framework}, similar to a VAE, we utilize a posterior neural network $g$ to parameterize each variational factor. Subsequently, applying the SGVB estimator, the evidence lower bound $\mathcal{L}_{ELBO}(y|x)$ in Equation (\ref{eq6}) can be reformulated as:
\begin{align}
    \mathcal{L}_{\text{ELBO}}(y|x)&= \mathbb{E}_{q(h,z,c,w|y,x)}[\log{\frac{p(y,h,z,c,w|x)}{q(h,z,c,w|y,x)}}] \\
    &= \mathbb{E}_{q(h,z,c,w|y,x)}[\log{\frac{p(y|x,h,z,c,w)p(h,z,c,w|x)}{q(h,z,c,w|y,x)}}] \\
    &= \mathbb{E}_{q(h,z,c,w|y,x)}[\log{p(y|x,h,z,c,w)}] \label{eq14} \\ 
    &\quad - KL[q(h,z,c,w|y,x)||p(h,z,c,w|x)] \nonumber
\end{align}
where the first term denotes the expected log-likelihood, measuring the construction ability from the pixelated trajectory and road network to a pixelated path. The second term represents the Kullback-Leibler divergence between variational prior and posterior, regularizing appropriate clusters and latent embedding for the map-matching scenario. Moreover, an MSE auxiliary loss is also utilized to guide the modeling of uncertainty characteristics. It leverages statistical information like average error distance and variance extracted from the ground-truth path.

\begin{algorithm}[t]
\caption{Constraint Path-Finding Algorithm}
\begin{algorithmic}[1]
\REQUIRE Candidate road set $S_y$, Road network $G(V,E)$, cost threshold $C'$, start road $e_1$, end road $e_n$
\STATE Initialize a priority queue $Q \gets \varnothing$, a result set $S$.
\STATE Generate a label $l_0: \mathcal{P}_0=e_1 (\text{cost}:0,\text{length}:0)$.
\STATE Insert $l_0$ into $Q$.
\WHILE{$Q \neq \varnothing$}
\STATE Pop a candidate path $\mathcal{P}_i \in Q$ with lowest cost.
\IF{the end road is reached}
    \STATE Insert $\mathcal{P}_i$ into $S$.
    \STATE continue
\ENDIF
\FOR{each neighbor road segment $e_j$}
\IF{$\text{Cost}(\mathcal{P}_i \cup e_j)\leq C'$}
    \STATE Generate a label $l': \mathcal{P}_i \cup e_j (\text{Cost}(\mathcal{P}_i \cup. e_j),\text{length}(\mathcal{P}_i \cup e_j))$.
    \IF{$l'$ is not dominated by any label in $Q$}
        \STATE Insert $l'$ into $Q$.
        \STATE Remove any dominated labels from $Q$.
    \ENDIF
\ENDIF
\ENDFOR
\ENDWHILE
\STATE Select resulting path $\mathcal{P}_{out}$ with lowest cost. If costs are equal, we select the path with the shortest length.
\ENSURE Resulting Path $\mathcal{P}_{out}$
\end{algorithmic}
\label{alg1}
\end{algorithm}
\subsection{Constraint Path-Finding}
After obtaining the calibrated trajectory at the pixel level, we convert it to a road ID sequence. While calibrated trajectories roughly pinpoint user positions, they lack awareness of connectivity between road segments and turn restrictions at junctions. Sometimes, they even produce discontinuous trajectories with blank spots or conflicting localization due to the inherent unpredictability of generative neural networks. Inspired by \cite{constraint1,constraint2,constraint3}, we propose a constraint path-finding algorithm that adheres to the topological feasibility. This algorithm searches for a path that closely follows the calibrated trajectory guidance while proactively employing a shortest-length preference to avoid matching detours when fuzzy localization occurs in the calibrated trajectory.

Formally, given a pixelated path $y$ comprising a set of pixel cells, each corresponding to a grid coordinate $(lon, lat)$ and a grid size $r$, we define our approach within the road network $G(V,E)$. If the endpoints $v_s$ and $v_e$ of a road segment $e_i \in E$ fall within the pixel cells of $y$, that road segment is included in the candidate road set $S_y$. For a path $\mathcal{P} = e_1, e_2, \dots, e_{|\mathcal{P}|}$, we define the deviation cost function as:
\begin{equation}
\text{Cost}(\mathcal{P}) = \sum_{i=1}^{|\mathcal{P}|} d(e_i)
\end{equation}
\begin{equation} \label{eq16}
d(e_i) = \begin{cases} 0, & \text{if $e_i \in S_y$} \\
\text{length}(e_i), & \text{otherwise} \end{cases}
\end{equation}
Where $d_i$ is the deviation cost of road $e_i$, and $\text{length}(e_i)$ represents the road length. Our optimization objectives are twofold: 1) find a path ensuring the deviation loss $\text{Cost}(\mathcal{P})$ remains below a specified threshold $C'$ (e.g., within 3\% of the trajectory length), and 2) minimize the path length $\text{length}(\mathcal{P})$ to avoid unnecessary detours.

To achieve these objectives, we employ a label setting based algorithm for efficient optimal path finding. We initialize the start and end road segments $e_1$ and $e_n$ as those corresponding to the endpoints of the pixelated path, which are closest to the first and last cellular trajectory points $p_1$ and $p_{|\mathcal{T}|}$, respectively. Algorithm \ref{alg1} details the constraint path-finding procedure. The algorithm begins its exploration from the start road segment $e_1$, expanding all topologically reachable paths and calculating their cost and length while satisfying $\text{Cost}(\mathcal{P})\leq C'$. For two different candidate paths $\mathcal{P}_1 = e_1, \dots, e_i, e_k$ and $\mathcal{P}_2 = e_1, \dots, e_j, e_k$ reaching the same road segment $e_k$, path 1 is considered to dominate path 2 if:
\begin{equation}
\text{Cost}(\mathcal{P}_1)\leq \text{Cost}(\mathcal{P}_2) \text{ and } \text{length}(\mathcal{P}_1)\leq \text{length}(\mathcal{P}_2)
\end{equation}
In such cases, we prune the exploration of path 2 to improve search efficiency. During path exploration, we prioritize expanding candidate paths with the minimum cost to efficiently search for the optimal path. From the candidate paths reaching the target road segment $e_n$, we select the one with the minimum cost as the result path $\mathcal{P}_{res}$. In cases of equal costs, we choose the path with the shortest length. This joint optimization process efficiently finds the globally optimal path $\mathcal{P}_{out}$ that conforms to road network constraints while minimally deviating from pixelated guidance and maintaining the shortest length to avoid detours.

\section{EXPERIMENTS}
In this section, we conduct a comprehensive evaluation of our approach's zero-shot performance. Initially, we outline the experimental setup. Subsequently, we detail the evaluation metrics employed. Following this, we benchmark our approach against both traditional and learning-based map-matching methods, presenting key findings along with an analytical discussion addressing several pertinent questions:
\begin{itemize}
    \item \textbf{Q1}: How does the performance of our ZSMM approach compare to that of existing map-matching methods?
    \item \textbf{Q2}: How well does the ZSMM approach transfer in a zero-shot context?
    \item \textbf{Q3}: What impact does each component of our design have on the ZSMM's performance?
    \item \textbf{Q4}: How do hyper-parameters affect ZSMM's performance?
    \item \textbf{Q5}: How can the workings of ZSMM be intuitively explained?
\end{itemize}
\subsection{Experimental Setting}

\begin{table}
\begin{center}
\small
\setlength{\tabcolsep}{3pt}
\fontsize{8.5}{12}\selectfont
\caption{Dataset characteristics.}
\label{table:dataset}
\begin{tabular}{ ccccc } 
\hline
 & Hangzhou & Xiamen & Chengdu & Beijing \\
\hline
\makecell{Road Segments} & 92,913 & 64,828 & 84,621 & 142,350 \\ 
\hline
\makecell{Cellular Trajectory \\ points (million)} & 3.61 & 1.18 & 2.17 & 2.55 \\ 
\hline
\makecell{Cellular Trajectory \\ Points per Trajectory}  & 34 & 40 & 37 & 35 \\ 
\hline
\makecell{Average Cellular \\ Sampling Distance (m)}  & 730 & 650 & 748 & 686 \\ 
\hline
Intersections & 67,330 & 37,591 & 58,254 & 84,672 \\
\hline
Turn constraint proportion & 17\% & 9\% & 16\% & 22\% \\
\hline
Main road proportion & 26\% & 41\% & 23\% & 24\% \\
\hline
\end{tabular}
\end{center}
\end{table}
\subsubsection{Dataset description}
Our experiments encompass both real-world and synthetically generated datasets. Specifically, we draw from real cellular trajectory data gathered in Hangzhou and Xiamen, China, which were procured from a mobile communication operator. These datasets not only include the raw cellular trajectories but also the matched GPS sampling sequences that trace the same routes, providing the necessary ground truth crafted through a conventional HMM algorithm \cite{ST-matching}.
To complement real-world data, we synthesized additional datasets leveraging publicly accessible GPS trajectories from Didi Chuxing for Chengdu and Beijing. These synthetic trajectories were constructed based on the cellular tower coordinates sourced from the OpenCellid dataset. Road network data for every dataset was sourced from OpenStreetMap. The core attributes of the datasets are outlined in Table \ref{table:dataset}. This diverse collection of datasets provides a robust platform for evaluating our map-matching approach across different geographic regions and varying data densities. 

\subsubsection{Parameter setup}
Through experimental comparison, we set the number of clusters to 32. The CNN contains 3 layers with 3$\times$3 kernels and the transformer contains 2 layers with 8 attention heads. The normalized pixel image size is $224 \times 224$. The dimensions of the hidden features $h_s$ and $z$ are both 1024. The sampled top-$k$ clustering number is 3. All trainable parameters are optimized using Adam ($\beta_1=0.9$, $\beta_2=0.999$) based on the training set, with hyperparameters selected on the validation set. The initial learning rate is $1 \times 10^{-5}$, with a cosine learning rate scheduler and 1000-step warmup. For constraint path-finding, we set the cost threshold as 3\% of the cellular trajectory length.

\subsubsection{Evaluation criteria}
Following previous work \cite{DMM,If-matching}, we evaluate the map-matching quality using Precision and Recall, measured by comparing the ground-truth path and the resulting path at the road segment level. Specifically, Precision is defined as the ratio of correctly matched road segment length to total matched road segment length. Recall is the ratio of correctly matched road segment length to total ground truth road segment length. Additionally, we use average inference time (Time) to evaluate the efficiency, defined as the average running time in seconds for map-matching one cellular trajectory on a machine with Intel Xeon E5-2680 v4 CPU and NVIDIA Tesla H800 GPU.

\begin{table*}[t]
\begin{center}
\normalsize
\setlength{\tabcolsep}{3pt}
\caption{Overall Performance. Underlined values indicate the best performance among baseline methods for each metric and dataset. Bold values highlight the best overall performance, achieved by our ZSMM method.}
\label{table:performance}
\begin{tabular}{cccccccccccccc} 
\hline
\textbf{Dataset} & \multicolumn{3}{c}{\textbf{Hangzhou}} &  \multicolumn{3}{c}{\textbf{Xiamen}} & \multicolumn{3}{c}{\textbf{Chengdu}} & \multicolumn{3}{c}{\textbf{Beijing}} \\
\textbf{Metric} & \textbf{Precision} & \textbf{Recall} & \textbf{Time} & \textbf{Precision} & \textbf{Recall} & \textbf{Time} & \textbf{Precision} & \textbf{Recall} & \textbf{Time} & \textbf{Precision} & \textbf{Recall} & \textbf{Time}  \\
\hline
\multicolumn{13}{c}{\textbf{Traditional heuristic-based map-matching methods}} \\
\hline
SNet \cite{Snapnet} & 0.446 & 0.555 & 0.034 & 0.475 & 0.565 & 0.041 & 0.463 & 0.560 & 0.035 & 0.457 & 0.558 & \underline{0.034} \\ 
THMM \cite{THMM} & 0.461 & 0.562 & 0.041 & 0.486 & 0.583 & 0.045 & 0.477 & 0.580 & 0.040 & 0.469 & 0.585 & 0.039 \\ 
\hline
\multicolumn{13}{c}{\textbf{Learning-based map-matching methods}} \\
\hline
DeepMM \cite{DeepMM} & 0.446 & 0.544 & 0.951 & 0.478 & 0.568 & 1.284 & 0.468 & 0.563 & 0.903 & 0.462 & 0.563 & 0.983 \\ 
TransMM \cite{transformerMM} & 0.455 & 0.552 & 1.667 & 0.483 & 0.577 & 1.857 & 0.480 & 0.588 & 1.475 & 0.467 & 0.572 & 1.925 \\ 
DMM \cite{DMM} & 0.467 & 0.566 & 0.853 & 0.489 & 0.594 & 0.916 & 0.485 & 0.607 & 0.874 & 0.480 & 0.581 & 0.966 \\ 
GraphMM \cite{liu2023graphmm} & 0.508 & 0.584 & 1.431 & 0.517 & 0.650 & 1.449 & 0.505 & 0.611 & 1.108 & 0.473 & 0.592 & 1.165 \\ 
LHMM \cite{LHMM} &  \underline{0.516} &  \underline{0.613} &  \underline{0.032} &  \underline{0.547} &  \underline{0.667} &  \underline{0.037} &  \underline{0.527} &  \underline{0.620} &  \underline{0.033} &  \underline{0.481} &  \underline{0.597} &  0.039 \\ 
\hline
\multicolumn{13}{c}{\textbf{Our method}} \\
\hline
\textbf{ZSMM} & \textbf{0.603} & \textbf{0.686} & \textbf{0.024} & \textbf{0.617} & \textbf{0.722} & \textbf{0.024} & \textbf{0.615} & \textbf{0.716} & \textbf{0.023} & \textbf{0.608} & \textbf{0.696} & \textbf{0.025} \\ 
\textbf{Improved} & \textbf{16.8\%} & \textbf{11.9\%} & \textbf{25.0\%} & \textbf{12.8\%} & \textbf{8.2\%} & \textbf{35.1\%} & \textbf{16.7\%} & \textbf{15.5\%} & \textbf{30.3\%} & \textbf{26.4\%} & \textbf{16.6\%} & \textbf{26.4\%} \\ 
\hline
\end{tabular}
\end{center}
\end{table*}
\subsubsection{Baselines}
We consider the following methods as baselines to compare. 
The traditional methods for CTMM include:
\begin{itemize}
    \item \textbf{SnapNet (SNet) \cite{Snapnet}.} It integrates digital map hints and a number of heuristics into the HMM process.
    \item \textbf{THMM \cite{THMM}.} It is an HMM-based model, which incorporates geometric, topological, and probabilistic characteristics to search for a reasonable path.
\end{itemize}
The learning-based methods for CTMM include:
\begin{itemize}
    \item \textbf{DeepMM \cite{DeepMM}.} It takes LSTM-based seq2seq and attention models to model a learnable map-matching function.
    \item \textbf{TransMM \cite{transformerMM}.} It replaces LSTM with Transformer in the seq2seq model.
    \item \textbf{DMM \cite{DMM}.} It adopts a reinforcement learning component to enhance the seq2seq map-matcher.
    \item \textbf{GraphMM \cite{liu2023graphmm}.} It adopts a graph neural network to extract inter-trajectory and trajectory-road correlation.
    \item \textbf{LHMM \cite{LHMM}.} It neuralized the observation and transition probabilities of HMM to model trajectory-road correlations.
\end{itemize}

\subsection{(Q1) Overall Performance}
Table \ref{table:performance} exhibits the results of our method’s performance compared with traditional HMM-based and learning-based methods on both accuracy and efficiency.

\textbf{Accuracy.} We observe that learning-based methods significantly outperform traditional HMM-based approaches. For instance, the neuralized HMM method LHMM achieves 0.516 precision and 0.613 recall, compared to conventional heuristic HMMs with 0.461 precision and 0.562 recall, respectively. This demonstrates the superiority of learning-based methods in capturing matching patterns between cell towers and roads by overcoming high positioning errors through data statistics. Under lower sampling rates, the precision of all models decreases, indicating greater challenges posed by higher localization uncertainty. Our proposed ZSMM outperforms previous methods on all four datasets, showcasing its effectiveness. For example, on the Hangzhou dataset, ZSMM achieves a precision of 0.603 and recall of 0.686, improving over the conventional HMM approach THMM by 30.8\% and the previously best learning-based method LHMM by 16.8\% in terms of precision and recall. Similar improvements are observed on the Xiamen dataset. The paradigm shift from ID embedding modeling to pixelated map-matching not only preserves but enhances accuracy by more robustly localizing user positions despite high positioning errors.

\textbf{Efficiency.} We observe that HMM-based methods are faster than seq2seq methods for map-matching. For example, LHMM processes a trajectory in 0.032 seconds, outpacing DMM at 0.853 seconds per trajectory. This speed advantage is attributed to HMM's parallelizable observation and transition probabilities calculation, whereas seq2seq methods are limited by the weak parallelism of RNN inference. On the other hand, our proposed ZSMM achieves the fastest matching speed of 0.024 seconds per trajectory, up to 35\% faster than LHMM on the Xiamen dataset. This superior efficiency stems from several factors: ZSMM's lightweight neural network computations, parallelizable VAE architecture with three separate computation branches; and constraint path-finding process is enhanced by label-based pruning, requiring fewer shortest-path computations than the Viterbi algorithm of HMM. Moreover, the pixelated path generation part remains constantly time-cost regardless of road network complexity, maintaining high efficiency across different scales.

\begin{table}[t]
\normalsize
\begin{center}
\caption{Zero-shot Performance.}
\label{table:zeroshot}
\begin{tabular}{ ccccc} 
\hline
\textbf{Dataset} & \multicolumn{2}{c}{\textbf{Xiamen}} &  \multicolumn{2}{c}{\textbf{Beijing}} \\
\textbf{Metric} & \textbf{Precision} & \textbf{Recall} & \textbf{Precision} & \textbf{Recall}  \\
\hline
THMM & 0.486 & 0.583 & 0.469 & 0.585 \\
DMM (HZ) & 0 & 0 & 0 & 0 \\
LHMM (HZ) & 0 & 0 & 0 & 0 \\
\hline
ZSMM (HZ) & 0.614 & 0.715 & 0.606 & 0.691 \\
ZSMM (CD) & 0.610 & 0.713 & 0.605 & 0.690 \\
\hline
\end{tabular}
\end{center}
\vspace{-0.3cm}
\end{table}

\subsection{(Q2) Zero-shot Performance}
We further evaluate our model's zero-shot map-matching performance on unseen regions in Table \ref{table:zeroshot}. To investigate the transferability of ZSMM, we designed several variants: DMM (HZ), LHMM (HZ), ZSMM (HZ), and ZSMM (CD), where HZ and CD represent models trained on Hangzhou and Chengdu datasets respectively. Our observations reveal that the heuristic method THMM shows slightly degraded performance on Beijing, which has more complex road networks and different cell tower distributions. The learning-based methods that rely on ID-embeddings, such as LHMM and DMM, completely fail to transfer across cities, yielding 0 precision and recall. These methods encode ID-based tower-road associations rather than generalizable patterns, requiring additional training data to function properly in new environments. 

In contrast, ZSMM demonstrates effective zero-shot transferability, maintaining precision around 0.614 and recall of 0.715 with minimal performance drop of less than 2\% compared to in-distribution testing. This robust transfer capability is attributed to ZSMM's modeling of transferable geospatial relationships between cellular trajectories and road networks in the pixel space, effectively learning region-agnostic matching patterns. Additionally, the GMM-based expert selection mechanism enables the model to dynamically identify and apply suitable experts to customize matching strategies for target regions with similar characteristics to those seen during training, further enhancing adaptability to new urban environments.

\begin{table}[t]
\normalsize
\begin{center}
\caption{Ablation Results.}
\label{table:ablation}
\begin{tabular}{ ccccc} 
\hline
\textbf{Dataset} & \multicolumn{2}{c}{\textbf{Hangzhou}} &  \multicolumn{2}{c}{\textbf{Beijing}} \\
\textbf{Metric} & \textbf{Precision} & \textbf{Recall} & \textbf{Precision} & \textbf{Recall}  \\
\hline
ZSMM (HZ) & 0.603 & 0.686 & 0.606 & 0.691 \\
ZSMM-T (HZ) & 0.583 & 0.669 & 0.587 & 0.674 \\
ZSMM-S (HZ) & 0.587 & 0.663 & 0.579 & 0.678 \\
ZSMM-E (HZ) & 0.576 & 0.657 & 0.558 & 0.653 \\
ZSMM-C (HZ) & 0.584 & 0.668 & 0.568 & 0.662 \\
\hline
\end{tabular}
\end{center}
\end{table}
\subsection{(Q3) Ablation Results}
In this part, we design a series of variants to examine the effects of key components in ZSMM. We train the models on the Hangzhou dataset and test on both Hangzhou and Beijing datasets.
\begin{itemize}
    \item ZSMM-T (HZ): Remove the temporal-aware network.
    \item ZSMM-S (HZ): Remove the spatial-aware network.
    \item ZSMM-E (HZ): Remove the GMM-based clustering and adaptive experts.
    \item ZSMM-C (HZ): Replace the constraint path-finding with a standard Dijkstra path-finding.
\end{itemize}
As depicted in Table \ref{table:ablation}, from the results of ZSMM-T (HZ), we observe that the sequential information is an effective complement in modeling long-range global dependencies, improving precision from 0.583 to 0.603 and recall from 0.669 to 0.686 in Hangzhou. According to ZSMM-S (HZ), explicitly modeling trajectory uncertainty characteristics such as offset distances helps overcome high positioning errors, with precision improved from 0.587 to 0.603 and recall from 0.663 to 0.686 in the same region. For ZSMM-E (HZ), we see a significant enhancement in transferability, where removing the localized experts leads to a remarkable drop in precision from 0.606 to 0.558 on the Beijing dataset. This indicates that using suitable experts can allow reasonable customization strategies tailored for the target region. 

Even when the train and test sets belong to the same Hangzhou city, ZSMM-E (HZ) shows that adaptive experts augment transferability in different regions, improving precision from 0.576 to 0.603. ZSMM-C (HZ) highlights the importance of constraint path-finding. Standard Dijkstra path-finding decreases precision in both Hangzhou and Beijing, showing a decline of approximately 2\% in precision and recall across both datasets. This demonstrates that our constraint path-finding algorithm more robustly balances trajectory guidance with road network constraints, leading to more accurate map-matching results. Overall, these ablation studies confirm that each component of ZSMM contributes significantly to its superior performance and zero-shot transferability across different urban environments.

\begin{figure}[t]
\centering
\begin{minipage}[t]{0.23\textwidth}
\begin{tikzpicture}[scale=0.47]
    \begin{axis}[
        xlabel=Cluster number,
        ylabel=Precision,
        y label style={at={(-0.05,0.5)}},
        ymajorgrids=true,
        font=\Large,
        ]
        \addplot+[color=red,line width=1.5pt,mark size=2pt] coordinates {
            (1,0.538)
            (8,0.587)
            (16,0.600)
            (32,0.603)
            (64,0.597)
            (128,0.595)
        };
    \end{axis}
\end{tikzpicture}
\caption{Impact of cluster number.}
\label{fig:clustering number}
\end{minipage}
\begin{minipage}[t]{0.23\textwidth}
\begin{tikzpicture}[scale=0.47]
    \begin{axis}[
        xlabel=Trajectory number (k),
        ylabel=Precision,
        y label style={at={(-0.05,0.5)}},
        ymajorgrids=true,
        scaled ticks = false,
        font=\Large,
        ]
        \addplot+ [color=blue,line width=1.5pt,mark size=2pt] coordinates {
            (3,0.513)
            (6,0.560)
            (9,0.573)
            (12,0.584)
            (15,0.587)
            (18,0.595)
            (21,0.599)
            (24,0.603)
        };
    \end{axis}
\end{tikzpicture}
\caption{Impact of data scales.}
\label{fig:data scales}
\end{minipage}
\end{figure}
\subsection{(Q4) Parametric Analysis}
In this section, we conduct a comprehensive investigation into the impact of key parameters on ZSMM performance, focusing on cluster numbers, training data scale, image resolution, and sampling frequency.

\subsubsection{Impact of Cluster Number}
The number of clusters is a critical hyperparameter that determines the regional categories and corresponding experts, directly affecting the model's adaptability to varied geographic contexts. We conducted experiments with cluster numbers ranging from 1 to 128 to understand how the granularity of region-specific expertise influences map-matching performance. A single cluster represents a global expert without regional specialization, while larger numbers create increasingly fine-grained regional specialists.

As shown in Figure \ref{fig:clustering number}, with a single cluster representing a global expert, the model achieves only baseline performance with a precision of 0.538. When increasing to 8 and 16 clusters, we observe a sharp improvement in precision, reaching 0.587 and 0.600 respectively. The performance peaks at 0.603 with 32 clusters, demonstrating the benefits of region-specific experts that can adapt to local road network characteristics and cell tower distributions. Further increasing the clusters to 64 and 128 results in a slight precision decline to 0.597 and 0.595, likely due to insufficient training data for numerous experts, leading to under-training of some regional specialists and subsequent performance degradation. This pattern suggests an optimal cluster number exists that balances specialization with sufficient training data per expert, which is particularly important for zero-shot transferability across regions.

\subsubsection{Impact of Data Scale}
Training data volume significantly influences the neural network's ability to generalize across diverse road networks and trajectory patterns. We systematically varied the training sample size from 3,000 to 24,000 trajectories to quantify how data availability affects model performance. This analysis is crucial for understanding the data requirements for effective zero-shot CTMM deployment in new regions.

Figure \ref{fig:data scales} illustrates the effect of varying training sample sizes on model precision. We observe a rapid improvement from 0.513 to 0.584 as samples increase from 3,000 to 12,000, representing an average precision gain of 0.023 per additional 3,000 samples. This substantial initial improvement indicates that the model requires a minimum threshold of diverse examples to learn transferable patterns between cellular trajectories and road networks. Beyond 12,000 samples, the improvement rate diminishes but remains positive, yielding a more modest average gain of 0.003 per 3,000 samples up to 24,000. The precision reaches 0.603 with 24,000 samples, and importantly, the upward trend continues without signs of convergence. This suggests potential for further performance improvements with larger datasets, indicating that our model architecture effectively utilizes additional training examples without saturation. The continued improvement with more data also confirms the model's ability to learn increasingly subtle and generalizable spatial relationships rather than merely memorizing specific trajectory-road pairings.

\begin{table}[t]
\centering
\normalsize
\setlength{\tabcolsep}{2pt}
\caption{Impact of image resolution on accuracy and running time for Hangzhou (HZ) and Xiamen (XM) datasets}
\label{tab:resolution-impact}
\begin{tabular}{lcccc}
\hline
Resolution & 112x112 & 224x224 & 384x384 & 512x512 \\
\hline
Accuracy (HZ) & 0.515 & 0.603 & 0.605 & 0.605 \\
Accuracy (XM) & 0.581 & 0.617 & 0.613 & 0.614 \\
Running Time (HZ) (s) & 0.018 & 0.024 & 0.026 & 0.033 \\
Running Time (XM) (s) & 0.017 & 0.023 & 0.025 & 0.031 \\
\hline
\end{tabular}
\end{table}

\subsubsection{Impact of Image Resolution}
The resolution of input images plays a crucial role in balancing performance and computational efficiency in our pixelated map-matching approach. We experimented with various image sizes ranging from 112×112 to 512×512 pixels, each representing a typical 1 kilometer trajectory. This parameter directly affects the granularity of spatial representation and the model's ability to distinguish closely positioned roads within dense networks.

Table \ref{tab:resolution-impact} presents our findings on accuracy and processing time across different resolutions. The results show substantial accuracy improvement from 0.515 to 0.603 when increasing resolution from 112×112 to 224×224 on the Hangzhou dataset. This significant gain indicates that the lower resolution fails to capture critical spatial details necessary for accurate road identification. However, further resolution increases yield diminishing returns, with accuracy plateauing at 0.605 for both 384×384 and 512×512 resolutions. This suggests that once a certain threshold of spatial detail is captured, additional resolution provides minimal benefit for the map-matching task. Regarding computational efficiency, running time increases progressively with higher resolutions, from 0.018 seconds at 112×112 to 0.033 seconds at 512×512. The 224×224 resolution offers an optimal balance, achieving high accuracy with a reasonable processing time of 0.024 seconds. At this resolution, each pixel approximates a 4-6 meter diameter area, adequately preserving spatial information without incurring excessive computational costs. Similar patterns were observed on the Xiamen dataset, confirming the generalizability of these findings across different urban environments.

\begin{table}[t]
\setlength{\tabcolsep}{3pt}
\centering
\caption{Effect of sampling interval on accuracy}
\label{tab:sampling-interval}
\begin{tabular}{lccccccc}
\hline
\makecell{Sampling Interval\\ (minutes)} & 0.2 & 0.4 & 0.6 & 0.8 & 1.0 & 1.2 & 1.4 \\
\hline
Accuracy on Hangzhou & 0.635 & 0.611 & 0.617 & 0.605 & 0.594 & 0.600 & 0.591 \\
Accuracy on Xiamen & 0.699 & 0.693 & 0.684 & 0.680 & 0.671 & 0.663 & 0.647 \\
Accuracy on Chengdu & 0.648 & 0.629 & 0.623 & 0.616 & 0.609 & 0.602 & 0.596 \\
Accuracy on Beijing & 0.642 & 0.624 & 0.618 & 0.611 & 0.602 & 0.595 & 0.590 \\
\hline
\end{tabular}
\end{table}

\subsubsection{Effect of Sampling Frequency}
Sampling frequency significantly affects map-matching performance, particularly for cellular trajectories with inherent positioning errors. We systematically varied the sampling interval from 0.2 to 1.4 minutes to investigate how temporal resolution impacts the model's ability to recover the true path. This analysis is particularly important for cellular data, where providers often impose sampling rate limitations due to privacy concerns and system constraints.

Our experiments across varying sampling intervals reveal consistent patterns across four datasets, as shown in Table \ref{tab:sampling-interval}. For the Hangzhou dataset, accuracy peaks at 0.635 with the shortest interval of 0.2 minutes and gradually decreases to 0.591 as the interval extends to 1.4 minutes, representing approximately a 7\% reduction in performance. Similarly, the Xiamen dataset shows accuracy declining from 0.699 to 0.647 across the same interval range, a decrease of about 7.4\%. This confirms that higher sampling frequencies provide more spatial information for accurate map-matching, especially in complex urban environments with dense road networks. However, we also note that the performance degradation is relatively gradual rather than precipitous, demonstrating our model's robustness to varying sampling rates. Furthermore, the minimal improvement between certain adjacent sampling rates suggests potential redundancy in excessively frequent sampling, indicating an efficient operating range for practical applications. This robustness to sampling frequency variations is particularly valuable for zero-shot deployment in regions where sampling constraints may differ from training data characteristics.

\begin{figure}[tbp]
\centering
\subfigure[Pixelated input]{
\begin{minipage}{0.46\linewidth}
\centering
\includegraphics[width=1.5in]{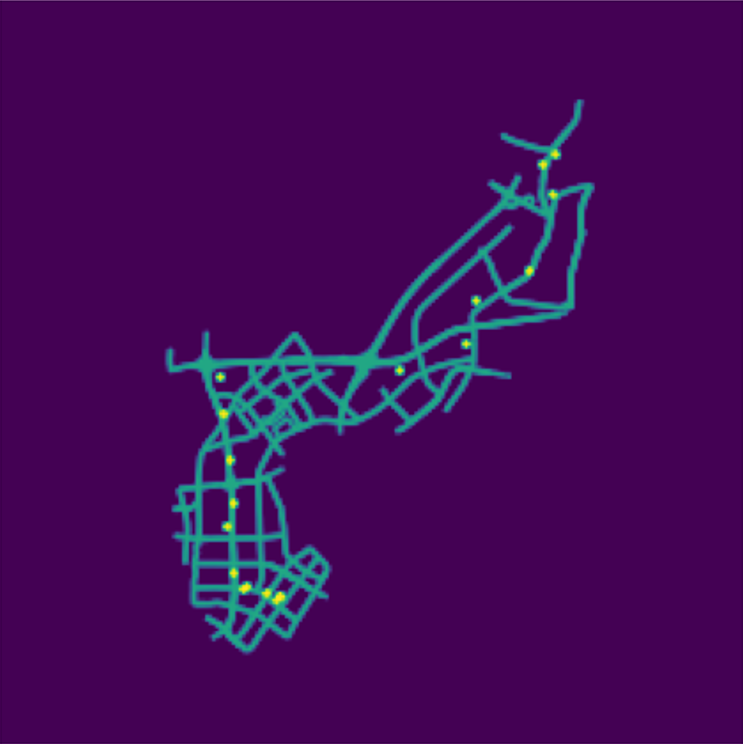}
\label{fig:visual1}
\end{minipage}
}
\subfigure[Calibrated trajectory]{
\begin{minipage}{0.46\linewidth}
\centering
\includegraphics[width=1.5in]{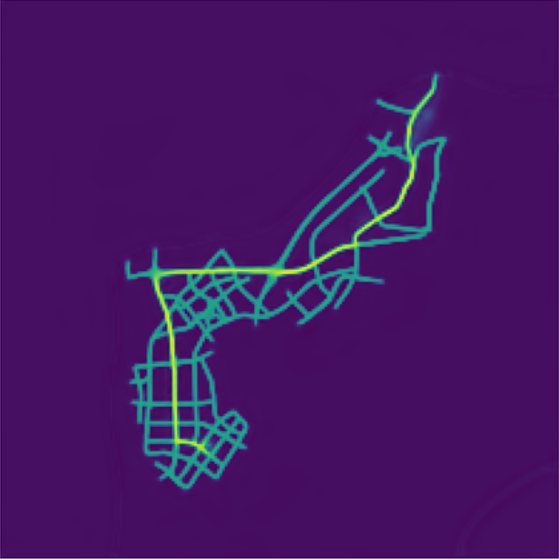}
\label{fig:visual2}
\end{minipage}%
}
\caption{Real cases of pixelated map-matcher.}
\end{figure} 

\subsection{(Q5) Visualization}
To provide qualitative insights into our pixelated map-matcher's performance, we present typical cases matched by ZSMM in Figure \ref{fig:visual1}. The figure illustrates the fundamental components of our visual approach: the input cellular trajectory (cyan) superimposed on the road network image (bright yellow) against a purple background. Figure \ref{fig:visual2} displays the corresponding calibrated trajectory output (yellow lines), which accurately identifies the user's geographic coordinates along the actual road segments.

The visualization reveals several important aspects of ZSMM's capabilities. Despite the highly sparse cellular trajectory with substantial positioning gaps between consecutive points, ZSMM effectively localizes users by leveraging contextual geospatial features encoded in the image representation. The complex road network structure in the urban area shown presents multiple potential path options at intersections, which our model navigates successfully. The calibrated trajectory correctly follows valid road segments even where the input trajectory appears to cut across non-road areas, showcasing the effectiveness of our constrained path-finding module. These visualizations confirm that the pixelated approach effectively captures the spatial relationships between trajectory points and road networks, allowing ZSMM to produce high-quality map-matching results even under challenging real-world conditions with limited cellular positioning accuracy.

\section{Complexity Analysis}
Understanding the time complexity of our approach is crucial for assessing its scalability and efficiency. Our method leverages lightweight CNNs and Transformers, which offer excellent parallelization capabilities.

Let $m$ be the number of roads, $d$ the hidden dimension, $W$ the image width, $K$ the CNN kernel size, and $C_{in}$ and $C_{out}$ the CNN input and output channel numbers, respectively. The time complexity of ZSMM can be expressed as: 
\[O(W^2K^2C_{in}C_{out} + W^2d + m^3)\]

This complexity can be broken down into three main components:
\begin{itemize}
    \item $O(W^2K^2C_{in}C_{out})$: Represents the CNN operations.
    \item $O(W^2d)$: Accounts for the Transformer operations.
    \item $O(m^3)$: Represents the worst-case cost for constrained path-finding.
\end{itemize}

In practice, the third term ($O(m^3)$) has negligible runtime impact. This is because paths deviating significantly from the calibrated trajectory are quickly pruned during the constrained path-finding process, substantially reducing the actual computational cost. A key advantage of our approach is that its computational cost is independent of trajectory length. This characteristic ensures stable running time across varying trajectory lengths and enables higher throughput under GPU parallel computing conditions. The efficiency of our model is further demonstrated by the running times observed in the image resolution experiments (Table \ref{tab:resolution-impact}). Even at the highest resolution of 512x512, the model completes processing in just 0.033 seconds, showcasing its practical applicability for real-time or near-real-time map-matching tasks.

For HMM-based models like LHMM, the complexity is $O(nk^2m \log m + Lmd^2)$, where $n$ is the number of trajectory points, $k$ is the number of candidate road segments for each observation point, $m$ is the number of roads, and $L$ represents the depth of the graph neural network. The $m \log m$ component reflects Dijkstra's algorithm complexity, which despite precomputation creates substantial memory overhead. LHMM's inference time grows linearly with trajectory length.

RNN-based models like DMM have a complexity of $O(nmd^2)$. DMM's inference time also scales linearly with trajectory length, and its sequential nature limits parallel processing capabilities on modern GPUs.

Compared to these methods, ZSMM is not affected by trajectory length and road network density, resulting in fast and stable inference speed. This makes our approach particularly suitable for real-time applications requiring consistent performance regardless of input complexity.

\section{Limitations}
Despite the promising performance of our zero-shot CTMM approach, several limitations warrant consideration. 

First, while ZSMM effectively handles static trajectory data, it faces challenges with streaming cellular trajectory data in real-time applications. The current pixelization approach processes complete trajectories, requiring modification for incremental processing of streaming data points. This limitation could be addressed by developing a sliding window mechanism that dynamically updates the pixelated representation as new data arrives, though this would increase computational complexity and potentially affect localization accuracy at trajectory boundaries.

Second, our model's performance heavily depends on the quality and coverage density of the road network data from OpenStreetMap. In regions with incomplete or outdated road information, the calibrated trajectories may align with non-existent or inaccurate road segments. This issue is particularly prominent in rapidly developing urban areas or regions with limited mapping resources. Future work could incorporate multi-source road network fusion and uncertainty quantification to handle inconsistencies between the calibrated trajectory and available road network data.

Third, while we demonstrate effective zero-shot transfer across four cities, the model may encounter challenges when generalizing to drastically different urban layouts and cell tower deployment strategies internationally. Cultural and infrastructural differences in city planning, such as grid-based versus radial layouts, or varying cellular network technologies and deployment densities, could affect the transferability of learned patterns. Additional evaluation across more diverse global urban environments would better establish the limits of ZSMM's zero-shot capabilities and reveal whether region-specific adaptations are necessary for optimal performance in significantly different urban contexts.

\section{Conclusion}
In this paper, we presented a pixel-based trajectory calibration assistant for zero-shot CTMM. Temporal-aware and spatial-aware networks are designed to eliminate high uncertainty by explicitly modeling sequential and deviation characteristics. Adaptive experts are empowered by the Gaussian mixture model through soft clustering to share knowledge in similar regions while personalizing strategies for distinct regions. A constrained path-finding algorithm robustly produces a valid road sequence result guided by calibrated trajectory and shortest length constraints. Extensive evaluations demonstrate that our model achieves state-of-the-art zero-shot performance.

\section*{ACKNOWLEDGMENT}
The authors would like to express their sincere gratitude to the anonymous reviewers for their insightful comments and valuable suggestions. This work is supported by the National Natural Science Foundation of China (Grant No. 62272334), the priority academic program development of Jiangsu higher education institutions.

\nocite{*} 
\bibliographystyle{IEEEtran}
\bibliography{references}  

\vspace{-1cm}
\begin{IEEEbiography}[{\includegraphics[width=1in,height=1.25in,clip,keepaspectratio]{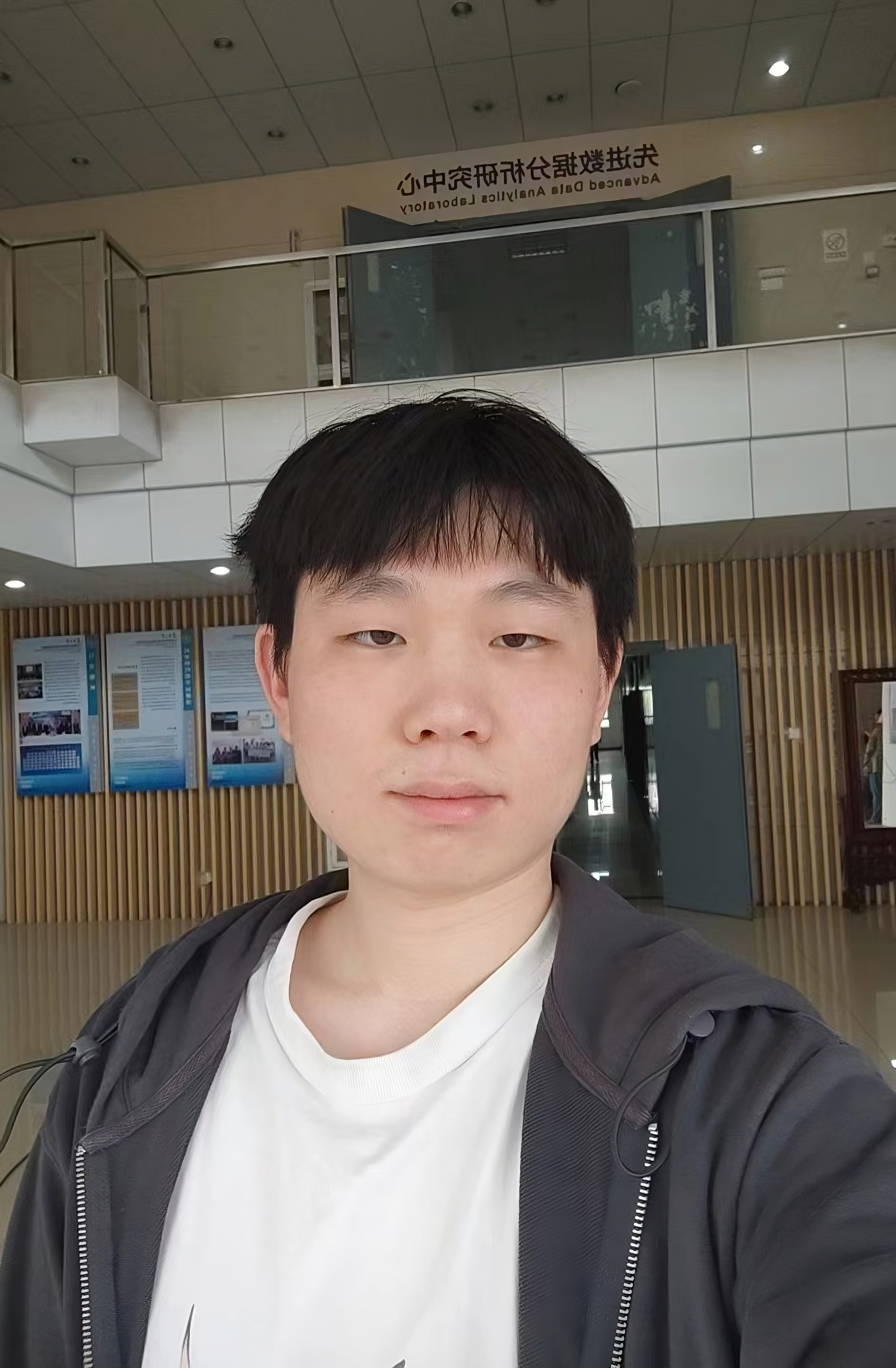}}]{Weijie Shi} received his Master's degree in 2024 from Soochow University. He is currently pursuing his Ph.D. degree at the Hong Kong University of Science and Technology (HKUST). His research interests include data management and spatiotemporal data mining.
\end{IEEEbiography}

\vspace{-0.5cm}
\begin{IEEEbiography}[{\includegraphics[width=1in,height=1.25in,clip,keepaspectratio]{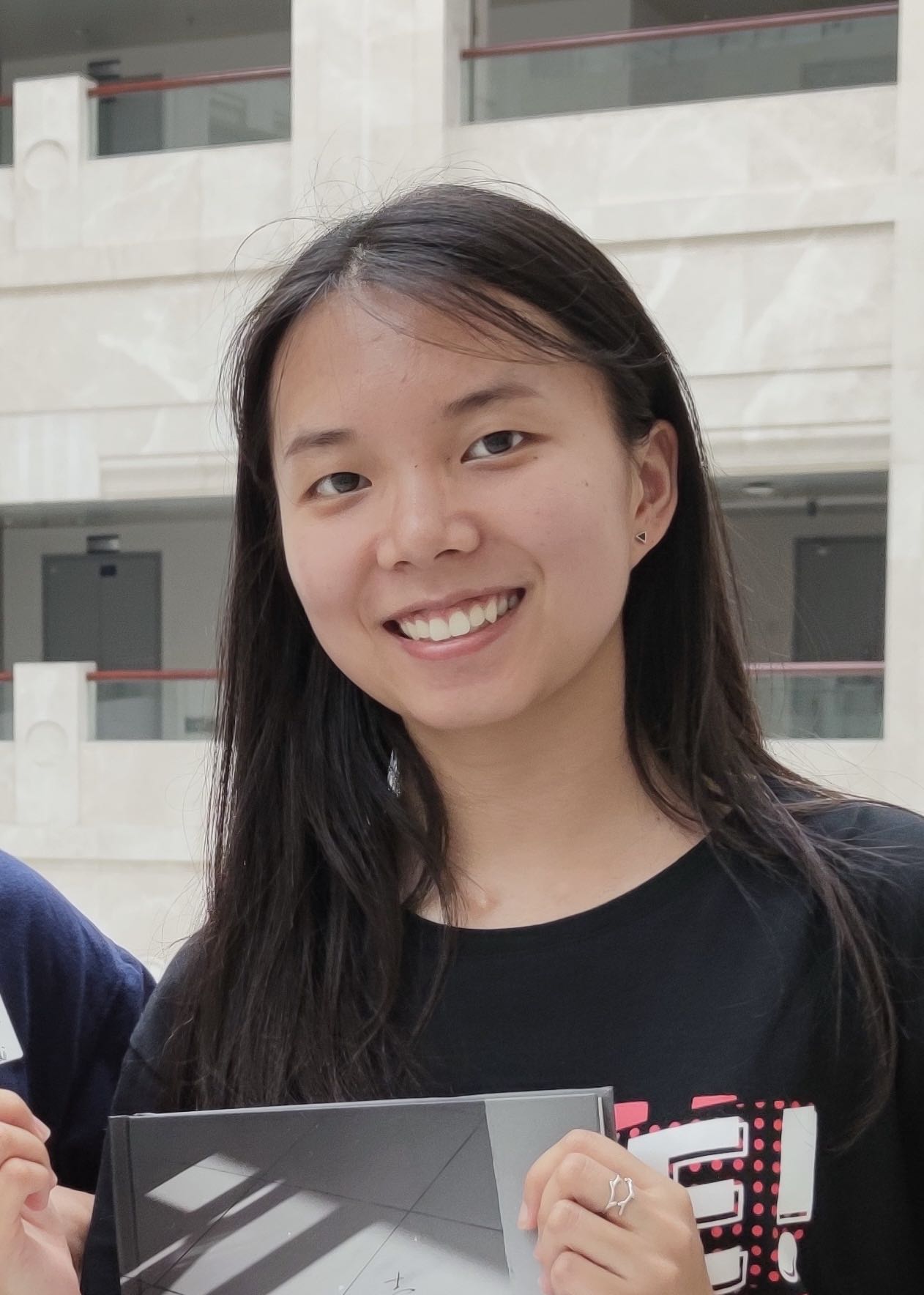}}]{Yue Cui} received the Bachelor's degree from the University of Electronic Science and Technology of China in 2020. She is currently pursuing her Ph.D. degree at the Hong Kong University of Science and Technology (HKUST). Her research focus is the interdisciplinary study of Data Science and Social Science, including reliability and governance. She is interested in developing frameworks for efficacy, equality, equity, and privacy.
\end{IEEEbiography}

\vspace{-0.5cm}
\begin{IEEEbiography}[{\includegraphics[width=1in,height=1.25in,clip,keepaspectratio]{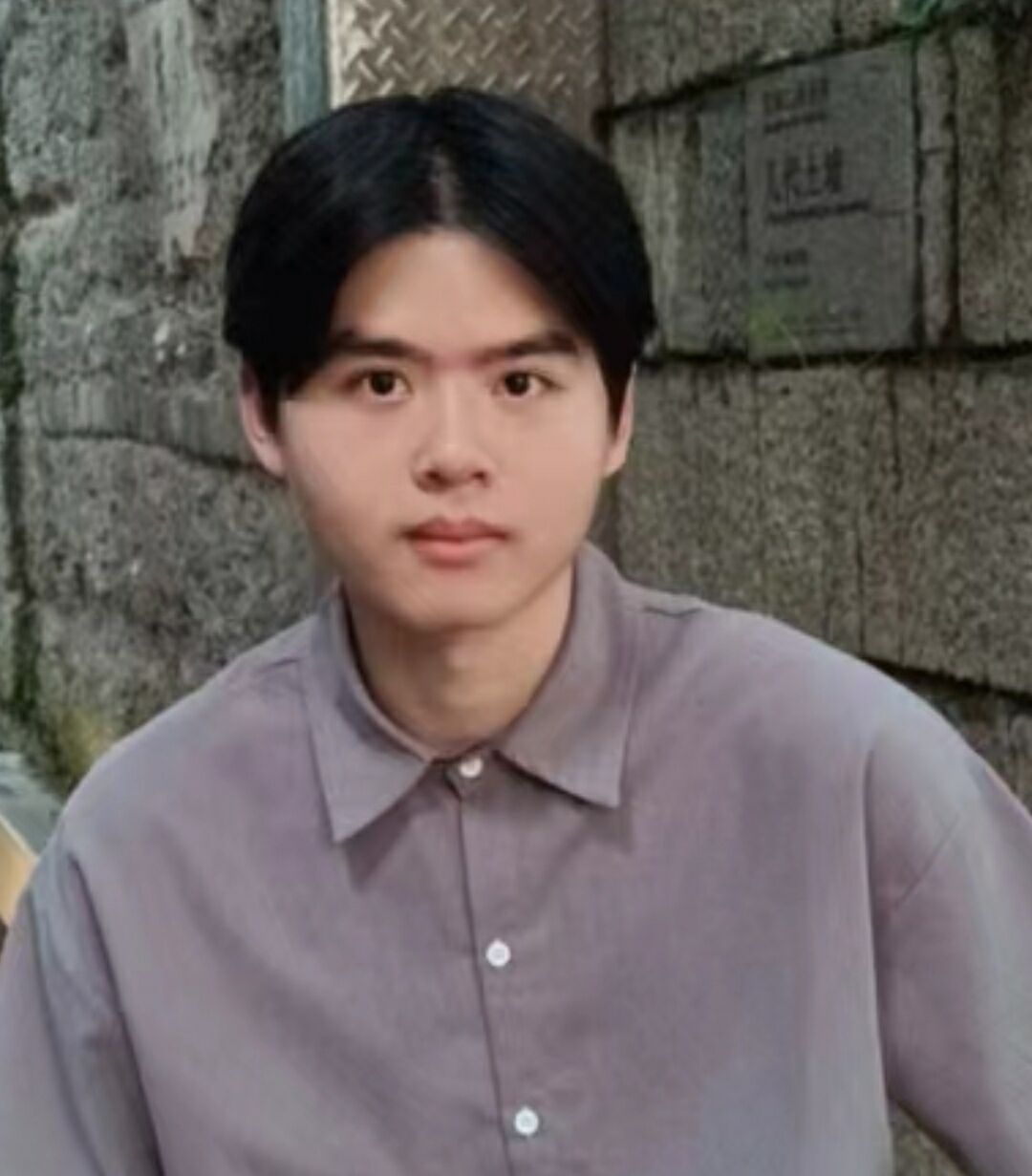}}]{Hao Chen} received his Master's degree from Sun Yat-sen University in 2024. He is currently working as an Engineer at Tencent Inc., Shenzhen, China. His research interests include multimodal deep learning and data mining.
\end{IEEEbiography}

\begin{IEEEbiography}[{\includegraphics[width=1in,height=1.25in,clip,keepaspectratio]{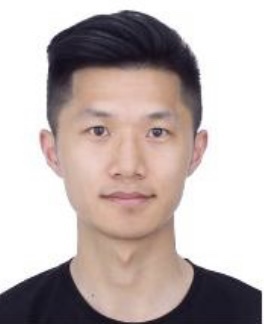}}]{Jiaming Li} received his Master's degree from Zhejiang University in 2021. He is currently working as an Engineer at ByteDance Inc., Hangzhou, China. His research interests include multimodal deep learning and data mining.
\end{IEEEbiography}

\begin{IEEEbiography}[{\includegraphics[width=1in,height=1.25in,clip,keepaspectratio]{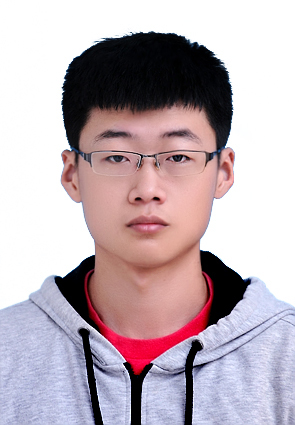}}]{Mengze Li} received his Ph.D. degree from Zhejiang University in 2024. He is currently a Postdoctoral Fellow at the Hong Kong University of Science and Technology. His research interests include multimodal deep learning and data mining.
\end{IEEEbiography}

\begin{IEEEbiography}[{\includegraphics[width=1in,height=1.25in,clip,keepaspectratio]{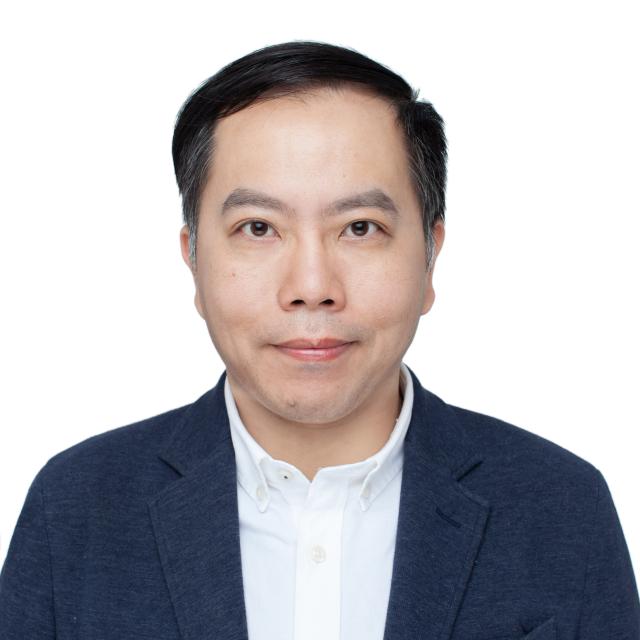}}]{Jia Zhu} received his Bachelor's and Master's degrees in Information Technology from Bond University, Australia in 2004 and 2005, respectively. He received his Ph.D. degree in Computer Science from the University of Queensland, Australia in February 2013. He is currently a Professor at the School of Education, Zhejiang Normal University, China. His research interests include educational large language models and spatiotemporal data analysis.
\end{IEEEbiography}

\begin{IEEEbiography}[{\includegraphics[width=1in,height=1.25in,clip,keepaspectratio]{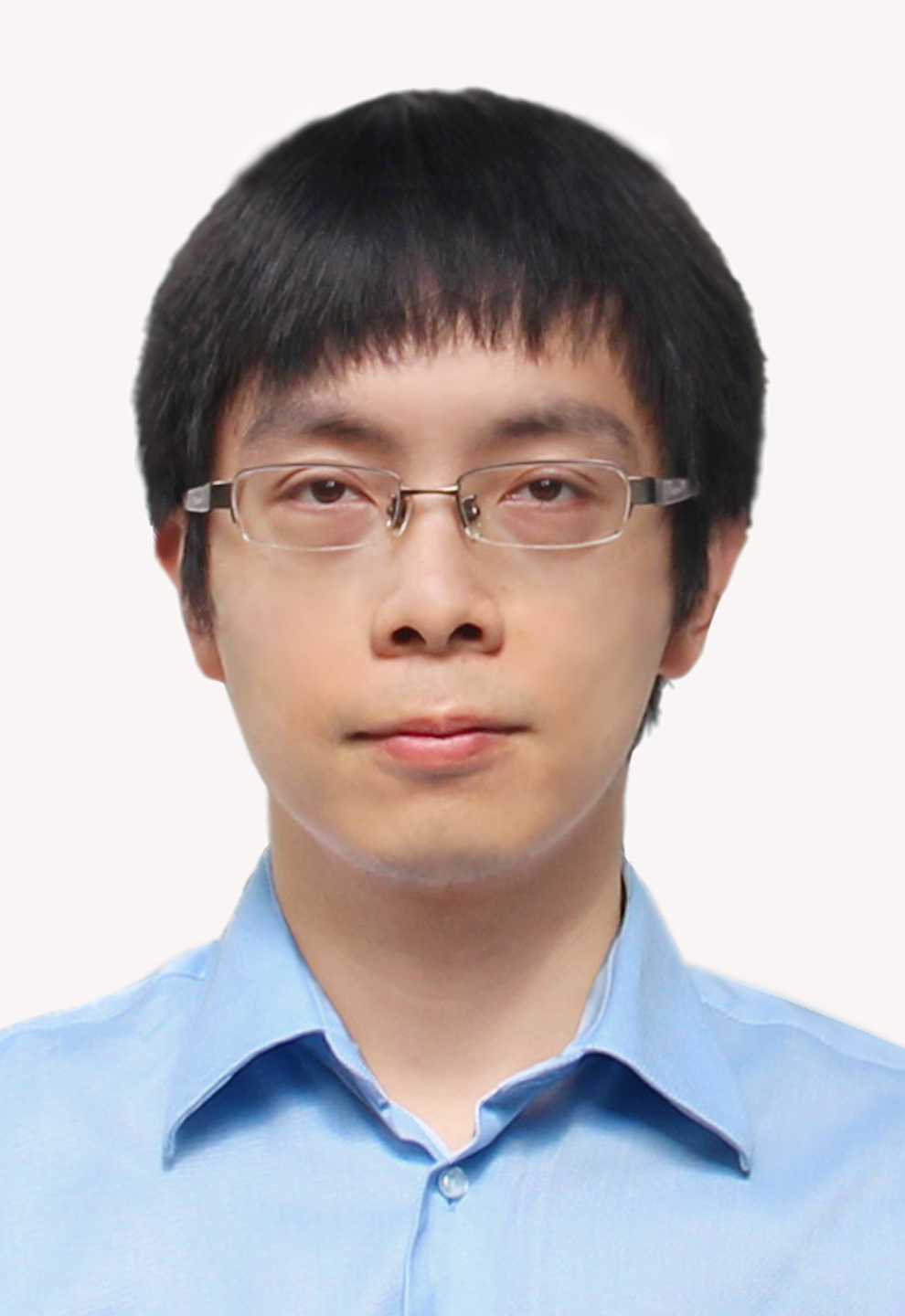}}]{Jiajie Xu} received his Ph.D. degree from Swinburne University of Technology, Australia in June 2011. He worked at the Institute of Software, Chinese Academy of Sciences from June 2011 to July 2013, before joining Soochow University in July 2013. He is currently a Professor at Soochow University, China. His research interests include databases, data mining, spatiotemporal computing, and intelligent recommendation systems.
\end{IEEEbiography}

\begin{IEEEbiography}[{\includegraphics[width=1in,height=1.25in,clip,keepaspectratio]{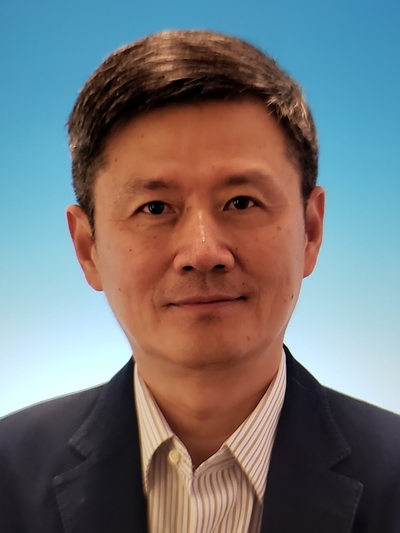}}]{Xiaofang Zhou} is Otto Poon Professor of Engineering and Chair Professor at HKUST. He received BSc and MSc degrees from Nanjing University and PhD from University of Queensland. Previously, he was a Professor at University of Queensland and Senior Research Scientist at CSIRO Australia. His research focuses on spatiotemporal data mining and big data analytics. He served as Program Chair for ICDE, CIKM, and PVLDB, and was Chair of IEEE Technical Committee on Data Engineering.
\end{IEEEbiography}

\end{document}